\documentclass[runningheads]{llncs}

\usepackage{amsfonts}
\usepackage{amsmath}
\usepackage{graphicx}
\usepackage{color}
\usepackage[caption=false]{subfig}
\usepackage{hyperref}

\newcommand\subjectto{\text{subject to}}
\newcommand\argmin{\text{argmin}}
\newcommand\ie{\textit{i.e.}\,}
\begin{document}

\title{Cost-Sensitive Best Subset Selection for Logistic Regression: A Mixed-Integer Conic Optimization Perspective}

\titlerunning{Cost-sensitive best subset selection for logistic regression}

\author{Ricardo Knauer \and
Erik Rodner}

\authorrunning{Knauer, R., and Rodner, E.}

\institute{KI-Werkstatt, University of Applied Sciences Berlin\\
\url{https://kiwerkstatt.htw-berlin.de}\\
\email{ricardo.knauer@htw-berlin.de}}

\maketitle

\begin{abstract}
A key challenge in machine learning is to design interpretable models that can reduce their inputs to the best subset for making transparent predictions, especially in the clinical domain. In this work, we propose a certifiably optimal feature selection procedure for logistic regression from a mixed-integer conic optimization perspective that can take an auxiliary cost to obtain features into account. Based on an extensive review of the literature, we carefully create a synthetic dataset generator for clinical prognostic model research. This allows us to systematically evaluate different heuristic and optimal cardinality- and budget-constrained feature selection procedures. The analysis shows key limitations of the methods for the low-data regime and when confronted with label noise. Our paper not only provides empirical recommendations for suitable methods and dataset designs, but also paves the way for future research in the area of meta-learning.
\end{abstract}

\section{Introduction}

Transparency is of central importance in machine learning, especially in the clinical setting \cite{din2022nrm,ec2021act}. The easier it is to comprehend a predictive model, the more likely it is to be trusted and applied in practice. Large feature sets decrease the comprehensibility and can therefore be problematic, which is particularly relevant for intrinsically interpretable models like logistic regression. Furthermore, it is often not reasonable to collect large numbers of features for predictive models in the first place. Consider a situation in which a physician would like to apply a decision support system that relies on a test battery of 5 questionnaires, 5 physical examination tests and 10 other assessments - measurement time constraints would make it impossible to utilize such a system in practice. Feature selection procedures therefore typically aim to reduce the feature set to the $k$ best predictors without sacrificing performance. The most intuitive way to select the best subset of $k$ features is to perform an exhaustive search over all possible feature combinations \cite{breiman2001statistical,guyon2003introduction}. However, this problem is NP-hard \cite{natarajan1995sparse}, which is why exhaustive enumeration is computationally intractable except for small predictor (sub)set sizes. There is a need for feature selection procedures that can find the best subset among all possible subsets for moderate (sub)set sizes, while taking into account the cost to collect predictors in practice \cite{guyon2003introduction,moons2019probast}, such as the limited amount of time for measuring them \cite{din2022nrm,steyerberg2019clinical}. In this work, we therefore focus on cost-sensitive best subset selection for an intrinsically interpretable machine learning model, logistic regression, which is often considered as a default for predictive modeling due to its simplicity \cite{christodoulou2019systematic,moons2015transparent,steyerberg2019clinical}.

The key contributions of our paper are as follows:
\begin{enumerate}
\item We propose a \textbf{cost-sensitive best subset selection} for logistic regression given a budget constraint, such as the available time to obtain predictors in practice. To the best of our knowledge and although there is a vast literature on feature selection~(Sect.~\ref{sec:relatedwork}), we provide the first mixed-integer conic optimization formulation for this problem (Sect.~\ref{csfs}).
\item We show how to generate \textbf{synthetic data for prognostic models} that can be used as a blueprint for future research, based on sensible parameter settings from the relevant clinical literature (Sect.~\ref{synth}).
\item This allows us to provide an \textbf{extensive evaluation of both heuristic and optimal cardinality- and budget-constrained feature selection procedures} with recommendations for the feature selection strategy, outcome events per variable, and
label noise level, and is also an important step towards meta-learning and foundation models in this domain (Sect.~\ref{experiments}).
\end{enumerate}

\section{Related work and preliminary definitions}
\label{sec:relatedwork}

In the following, we provide a brief overview about heuristics that have been proposed over the years to tackle the best subset selection problem without certificates of optimality. We then present contemporary approaches that solve the feature selection problem to provable optimality given an explicit cardinality constraint $k$ as well as approaches that consider the cost to measure predictors.

\subsubsection{Heuristic feature selection}

Heuristic feature selection procedures can be divided into three broad categories: filters, wrappers, and embedded methods \cite{guyon2003introduction}. Filters reduce the feature set without using the predictive model itself and are frequently based on the Pearson correlation coefficient \cite{guyon2004result} or univariable classifiers \cite{evans2022estimating}. In comparison to filters, wrappers employ the predictive model as a black box to score feature subsets. Prominent examples are forward or backward selection procedures. In forward selection, the most significant features are added to the predictive model; in backward selection, the most insignificant features are removed from the model \cite{abbott2014accuracy,van2021developing}. Both can be performed either once or stepwise, and are often combined in practice \cite{bakker2007spinal,dionne2011five,kennedy2006eight,kuijpers2006clinical}. Finally, embedded methods incorporate some form of subset selection into their training process. A well-known embedded method is L1-regularization which indirectly selects predictors by shrinking some of the model parameters to exactly zero \cite{tibshirani1996regression,wippert2017development}. A more direct way to induce sparsity is to penalize the number of nonzero parameters via L0-regularization \cite{bertsimas2019machine,dedieu2021learning}.

\subsubsection{Cardinality-constrained feature selection}

In practice, it is often desirable to explicitly constrain the number of nonzero parameters to a positive scalar $k$. Feature selection with such an explicit cardinality constraint has greatly benefited from recent advances in mathematical optimization \cite{bertsimas2021sparse,dedieu2021learning,deza2022safe,tamura2022feature}. Mathematical programming has not only made it possible to select the best subset of $k$ features for moderate (sub)set sizes, but also to deliver certificates of optimality for the feature selection process, at least for some predictive models (see \cite{tillmann2021cardinality} for a survey). 

We introduce the following notation. Let $\boldsymbol{X}\in\mathbb{R}^{M\times N}$ be the design matrix of $M$ training examples and $N$ features, $\boldsymbol{x}_m\in\mathbb{R}^N$ a single training example, $\boldsymbol{y}\in\{-1,1\}^M$ the labels, $y_m\in\{-1,1\}$ a single label, $\boldsymbol{\theta}\in\mathbb{R}^N$ the model parameters, $\theta_n\in\mathbb{R}$ a single parameter, $\theta_0\in\mathbb{R}$ the intercept, $\Vert\cdot\Vert_0\in\mathbb{R}_+$ the number of nonzero parameters, \ie, the L0 "norm", $\lambda\in\mathbb{R}_+$ a regularization hyperparameter for the squared L2 norm, and $k\in\mathbb{R}_+$ a sparsity hyperparameter for the L0 "norm". In \cite{bertsimas2021sparse}, the feature selection problem was solved to provable optimality for two predictive models, namely support vector classification with a hinge loss $\ell(y_m, \boldsymbol{\theta}^T\boldsymbol{x}_m+\theta_0) = \text{max}(0, 1 - y_m(\boldsymbol{\theta}^T\boldsymbol{x}_m+\theta_0))$ and logistic regression with a softplus loss $\ell(y_m, \boldsymbol{\theta}^T\boldsymbol{x}_m+\theta_0) = \log(1+\exp(-y_m(\boldsymbol{\theta}^T\boldsymbol{x}_m+\theta_0)))$, by means of the following optimization objective:

\abovedisplayskip0pt
\begin{equation}
\label{eqn:fs}
\begin{aligned}
\boldsymbol{\theta}^\ast, \theta_0^\ast 
 = \underset{\boldsymbol{\theta}, \theta_0}{\argmin} &\quad \sum_{m=1}^M \ell(y_m, \boldsymbol{\theta}^T\boldsymbol{x}_m+\theta_0) + \frac{\lambda}{2} \sum_{n=1}^N \theta_n^2 \\
\subjectto &\quad \Vert\boldsymbol{\theta}\Vert_0 \le k \quad.
\end{aligned}
\end{equation}

\noindent Using pure-integer programming with a cutting-plane algorithm, optimal solutions to eq.~\eqref{eqn:fs} could be obtained within minutes for feature set sizes up to $5000$ and subset sizes up to $15$. A drawback of the employed algorithm is its stochasticity, \ie, it is not guaranteed that successive runs yield similar results.
More recently, the cardinality-constrained feature selection problem \eqref{eqn:fs} was approached from a mixed-integer conic optimization perspective for logistic regression, and it was shown that optimal solutions can be found within minutes for features set sizes as large as $1000$ and subset sizes as large as $50$ \cite{deza2022safe}. In contrast to \cite{bertsimas2021sparse}, the mixed-integer conic solver in \cite{deza2022safe} is designed to be run-to-run deterministic \cite{aps2023mosek}, which makes the feature selection process more predictable. We extend the evaluation of \cite{deza2022safe} by assessing the cardinality-constrained conic optimization formulation with respect to the interpretability of the selection process (Sect.~\ref{experiments}), therefore asking whether this method is worth to use in practice.

\subsubsection{Budget-constrained feature selection}

In addition to heuristic and cardinality-constrained approaches to best subset selection, there have also been studies that considered the budget, especially the available time, to measure predictors in practice. In \cite{kuijpers2006prediction}, easily obtainable predictors were included in a logistic regression model first, before controlling for them and adding significant predictors that were harder to measure. In a patient-clinician encounter, asking questions during history-taking often takes less time than performing tests during physical examination, which is why predictors derived from physical examination were considered last. This hierarchical forward selection was followed by a stepwise backward selection. Similarly, features from history-taking were included and controlled for before features from physical examination in \cite{scheele2013course}.

Next to the aforementioned heuristics, a cost-sensitive best subset selection procedure for support vector classification was proposed in \cite{aytug2015feature}. In particular, the summed costs associated with the selected features were constrained to not exceed a predefined budget. Generalized Benders decomposition was used to allow the mixed-integer program to be solved for moderate feature (sub)set sizes in a reasonable amount of time. The procedure was then assessed on a range of synthetic and real-world problems, with each cost set to one and the budget set to $k$, \ie, with an explicit cardinality constraint as a special case of budget constraint (Sect.~\ref{csfs}). Best subsets could be selected within minutes on many of the problems, for instance with $N = 2000$ and $k = 17$ where similar approaches have struggled \cite{labbe2019mixed}. More recently, budget-constrained best subset selection for support vector classification using mixed-integer optimization was also evaluated for moderate (sub)set sizes with individual feature costs \cite{lee2020mixed}. No hyperparameter tuning was performed, which makes it difficult to put the experimental results into context with previous studies as predictive performance and computational time can vary significantly depending on the hyperparameter setting \cite{aytug2015feature,bertsimas2021sparse,deza2022safe,labbe2019mixed}.

Overall, there is an extensive literature for heuristic and mathematical optimization approaches to feature selection. Mathematical programming \cite{bertsimas2021sparse,dedieu2021learning,deza2022safe,tamura2022feature} can deliver certifiably optimal solutions for budget-constrained best subset selections, with cardinality-constrained best subset selections as a special case. With respect to logistic regression, a mixed-integer conic optimization perspective seems particularly appealing since it can deal with moderate feature (sub)set sizes while still being run-to-run deterministic \cite{deza2022safe,aps2023mosek}. To the best of our knowledge, we present the first optimal budget-constrained feature selection for logistic regression from a mixed-integer conic programming perspective (Sect.~\ref{csfs}) and systematically evaluate it in terms of the interpretability of the selection process (Sect.~\ref{experiments}).

\section{Best subset selection} \label{bss}

In this section, we describe two ways to approach optimal feature selection for logistic regression in case of moderate feature (sub)set sizes. We first review how to arrive at the best feature subset given a cardinality constraint. Based on that, we propose a new way to solve the best subset selection problem given a budget constraint, such as the available time to measure predictors in practice.

\subsection{Cardinality-constrained best subset selection} \label{ecc}

A cardinality-constrained formulation for best subset selection for logistic regression has already been introduced in eq.~\eqref{eqn:fs}:

\begin{equation}
\begin{aligned}
\boldsymbol{\theta}^\ast, \theta_0^\ast 
 = \underset{\boldsymbol{\theta}, \theta_0}{\argmin} &\quad \sum_{m=1}^M \log\left(1+\exp\left(-y_m(\boldsymbol{\theta}^T\boldsymbol{x}_m+\theta_0)\right)\right) + \frac{\lambda}{2} \sum_{n=1}^N \theta_n^2 \\
\subjectto &\quad \Vert\boldsymbol{\theta}\Vert_0 \le k \quad.
 \end{aligned}
\end{equation}

\noindent The first term of the optimization objective is known as the softplus (loss) function, the second term adds L2-regularization to prevent overfitting and make the model robust to perturbations in the data \cite{bertsimas2018characterization,bertsimas2019robust}, and the constraint limits the L0 "norm". However, explicitly constraining the number of nonzero parameters is not easy (in fact, it is NP-hard). To model our L2-regularized L0-constrained logistic regression, we choose a conic formulation similar to \cite{deza2022safe}. This allows us to restructure our optimization problem so that it can be solved within a reasonable amount of time in practice, even for moderate feature (sub)set sizes.

In conic programming, $\boldsymbol{\theta}$ is bounded by a $K$-dimensional convex cone $\mathcal{K}^K$, which is typically the Cartesian product of several lower-dimensional cones. A large number of optimization problems can be framed as conic programs. With a linear objective and linear constraints, for example, the nonnegative orthant $\mathbb{R}_+^N$ would allow us to solve linear programs \cite{ben2001lectures,boyd2004convex}. For our nonlinear objective, though, we use $2M$ exponential cones as well as $N$ rotated quadratic (or second-order) cones. The exponential cone is defined as follows \cite{aps2022mosek}:

\abovedisplayskip0pt
\begin{equation}
\mathcal{K}_{exp}=\{[\vartheta_1,\vartheta_2,\vartheta_3] \,|\, \vartheta_1 \ge \vartheta_2 e^{\vartheta_3 / \vartheta_2}, \vartheta_2 > 0\} \cup \{[\vartheta_1, 0, \vartheta_3] \,|\, \vartheta_1 \ge 0, \vartheta_3 \le 0\} \,.
\end{equation}

\noindent With this cone, we can model our softplus function using the auxiliary vectors $\boldsymbol{t}, \boldsymbol{u}, \boldsymbol{v}\in\mathbb{R}^M$:

\abovedisplayskip0pt
\begin{equation}
\begin{aligned}
&[u_m, 1, -y_m(\boldsymbol{\theta}^T\boldsymbol{x}_m+\theta_0)-t_m], [v_m, 1, -t_m]\in\mathcal{K}_{exp}, u_m + v_m \le 1 \\
&\Leftrightarrow t_m \ge \log(1+\exp(-y_m(\boldsymbol{\theta}^T\boldsymbol{x}_m+\theta_0))) \enspace\forall\,1 \leq m \leq M \enspace.
\end{aligned}
\end{equation}

\noindent The rotated quadratic cone is given by \cite{aps2022mosek}:

\abovedisplayskip0pt
\begin{equation}
\mathcal{Q}_R^K=\{ \boldsymbol{\vartheta}\in\mathbb{R}^K \,|\, \vartheta_1\vartheta_2 \ge \frac{1}{2}\Vert\boldsymbol{\vartheta}_{3:K}\Vert_2^2, \vartheta_1, \vartheta_2 \ge 0 \} \enspace.
\end{equation}

\noindent We use the rotated quadratic cone to represent our L2-regularization with the auxiliary vectors $\boldsymbol{z}\in\{0,1\}^N$ and $\boldsymbol{r}\in\mathbb{R}_+^N$. Additionally, the cone naturally allows us to set $\theta_n=0$ if $z_n=0$:

\abovedisplayskip0pt
\begin{equation}
[z_n, r_n, \theta_n]\in\mathcal{Q}_R^3 \Leftrightarrow z_n r_n \ge \frac{1}{2} \theta_n^2 \enspace\forall\,1 \leq n \leq N \enspace.
\end{equation}

\noindent Finally, we use an explicit cardinality constraint to limit the number of nonzero parameters:

\abovedisplayskip0pt
\begin{equation}
\label{eqn:ecc}
\sum_{n=1}^{N}{z_n \le k} \enspace .
\end{equation}

\noindent Our reformulated optimization problem therefore becomes:

\abovedisplayskip0pt
\begin{equation}
\label{eqn:ecc_long}
\begin{aligned}
\boldsymbol{\theta}^\ast, \theta_0^\ast = \underset{\boldsymbol{\theta}, \theta_0}{\argmin} &\quad \sum_{m=1}^M t_m + \lambda \sum_{n=1}^N r_n \\
\subjectto &\quad [u_m, 1, -y_m(\boldsymbol{\theta}^T\boldsymbol{x}_m+\theta_0)-t_m] \in\mathcal{K}_{exp} \enspace\forall\,1 \leq m \leq M\\
&\quad [v_m, 1, -t_m] \in\mathcal{K}_{exp} \enspace\forall\,1 \leq m \leq M\\
&\quad u_m + v_m \le 1 \enspace\forall\,1 \leq m \leq M\\
&\quad [z_n, r_n, \theta_n]\in\mathcal{Q}_R^3 \enspace\forall\,1 \leq n \leq N\\
&\quad \sum_{n=1}^{N}{z_n \le k}\\
&\quad \boldsymbol{\theta}\in\mathbb{R}^N, \boldsymbol{t},\boldsymbol{u},\boldsymbol{v}\in\mathbb{R}^M, \boldsymbol{r}\in\mathbb{R}_+^N, \boldsymbol{z}\in\{0,1\}^N \enspace .
\end{aligned}
\end{equation}

\subsection{Budget-constrained best subset selection} \label{csfs}

Cardinality-constrained best subset selection for logistic regression is already very useful to limit the number of candidate predictors. However, there are situations in which the cardinality is not the most suitable criterion to select features in practice, such as when two less costly predictors are preferred over one costly predictor. For instance, given a limited time budget to perform a diagnostic process, it may be more practical for a clinician to ask two quick questions rather than to perform one lengthy physical examination test. We therefore propose a cost-sensitive best subset selection procedure for logistic regression from a mixed-integer conic optimization perspective. To that end, we generalize eq.~\eqref{eqn:ecc} by using a cost vector $\boldsymbol{c}\in\mathbb{R}^N$ and a budget $b\in\mathbb{R}_+$ \cite{aytug2015feature,labbe2019mixed,lee2020mixed}:

\abovedisplayskip0pt
\begin{equation}
\label{eqn:csfs_long}
\sum_{n=1}^{N}{c_n z_n \le b} \enspace .
\end{equation}

\noindent The budget constraint forces the summed costs that are associated with the selected features to be less than or equal to our budget $b$. Note that this formulation can be regarded as a $0/1$ knapsack problem; like a burglar who aims to find the best subset of items that maximize the profit such that the selected items fit within the knapsack, we aim to find the best subset of features that maximize the log likelihood of the data such that the selected features fit within our budget. If $\boldsymbol{c} = \boldsymbol{1}$ and $b = k$, we recover the explicit cardinality constraint. Our budget-constrained best subset selection procedure is therefore more general than its cardinality-constrained counterpart. This allows us to capture situations in which $\boldsymbol{c} \neq \boldsymbol{1}$, \ie, individual feature costs can be different from one another, and increases the applicability of best subset selection.

\section{Data synthesis for prognostic models} \label{synth}

Machine learning models can be broadly deployed in a variety of settings, so getting a comprehensive picture about their performance is essential. Synthetic dataset generators are useful tools to systematically investigate predictive models under a diverse range of conditions and thus give recommendations about key performance requirements. They also provide an opportunity to pretrain complex machine learning models so that small real-world datasets can be used more efficiently \cite{hollmann2022tabpfn}. In the following, we describe our approach to synthetic data generation in the domain of prognostic model research. In particular, we derive parameter settings to generate sensible prognostic model data for multidimensional pain complaints, based on information from history-taking and physical examination. This setting is appealing because prognostic models appear to generally refine clinicians’ predictions of recovery in this domain \cite{abbott2014accuracy,hancock2009can,kennedy2006eight} and thus hold promise to make a large impact on clinical decision-making and outcomes.

Synthetic data can be generated through a variety of mechanisms. However, only few algorithms have served as a benchmark for a broad range of feature selection and classification methods \cite{guyon2004result,guyon2007competitive} or have been recently used in the context of best subset selection with mixed-integer programming \cite{tamura2022feature}. Therefore, we consider the algorithm that was designed to generate the MADELON dataset of the NIPS 2003 feature selection challenge to be a solid baseline for data synthesis \cite{guyon2003design}. In MADELON, training examples are grouped in multivariate normal clusters around vertices of a hypercube and randomly labeled. Informative features form the clusters; uninformative features add Gaussian noise. The training examples and labels can also be conceptualized to be sampled from specific instantiations of structural causal models (SCMs). Prior causal knowledge in the form of synthetic data sampled from SCMs has been recently used to meta-train complex deep learning architectures, \ie, transformers, for tabular classification problems, allowing them to very efficiently learn from only a limited amount of real-world (for instance clinical) data \cite{hollmann2022tabpfn}.

To derive sensible parameter settings for the MADELON algorithm, we scanned the prognostic model literature for multidimensional pain complaints, based on information from history-taking and physical examination, and chose the number of training examples and features according to the minimum and maximum values that we encountered. We thus varied the number of training examples between 82 \cite{evans2022estimating} and 1090 \cite{dionne2011five} and the number of features between 14 \cite{bakker2007spinal,van2021developing} and 53 \cite{evans2022estimating}. We also varied the label noise level between 0\% and a relatively small value of 5\% \cite{steyerberg2019clinical}  to simulate the effect of faulty missing label imputations, by randomly flipping the defined fraction of labels. This yielded 8 diverse synthetic datasets in total. We then made the classification on these datasets more challenging by two means. First, we added one “redundant” feature as a linear combination of informative features, which can arise in modeling when both questionnaire subscores and the summary score are included, for example \cite{evans2022estimating}. Second, it is quite common that classes are imbalanced in prognostic model research, which is why we added a class imbalance of 23\% / 77\% \cite{evans2022estimating}. The number of labels in the smaller class per feature, or outcome events per variable (EPV), thus roughly formed a geometric sequence from 0.36 to 17.91. We did not aim to investigate how to effectively engineer nonlinear features for our generalized linear logistic regression model, therefore we limited the number of clusters per class to one. Finally, as a preprocessing step, we ``min-max'' scaled each feature.

Each of these 8 datasets was used to assess our cardinality- and budget-constrained feature selection procedures. For our cardinality-constrained feature selection, we performed 4 runs with the number of informative features set to 2, 3, 4, or 5. For our budget-constrained feature selection, we performed 5 runs with the number of informative features set to 5 and each feature being assigned an integer cost between 1 and 10 sampled from a uniform distribution, \ie, $c_n \sim \mathcal{U}(1, 10)$ \cite{lee2020mixed}. For illustrative purposes, we assume that $c_n$ corresponds to the time in minutes that it takes to collect the feature in practice.

\section{Experiments} 
\label{experiments}

In this section, we describe the experimental setup to evaluate our cardinality- and budget-constrained feature selection procedures, including the mixed-integer conic optimization formulations (Sect.~\ref{bss}), on our 8 synthetic datasets (Sect.~\ref{synth}), and our cardinality-constrained best subset selection on real-world clinical data. We used MOSEK 10 \cite{aps2023mosek} from Julia 1.8.3 via the JuMP 1.4.0 interface \cite{dunning2017jump} for all experiments, with the default optimality gaps and no runtime limit. Finally, we present and discuss the results. In summary, interpretable results are only achieved at $EPV=17.91$, in line with other EPV recommendations of at least 10 \cite{heinze2018variable} to 15 \cite{harrell2015regression}, and without label noise for our optimal selection strategies. Please refer to the appendix for a more nuanced presentation.

\subsection{How does our approach perform compared to other models?} 
\label{baseline}

\subsubsection{Experimental setup}
Prior to our extensive evaluation on the 8 synthetic datasets, we conducted a baseline test for our mixed-integer conic optimization formulation on the Oxford Parkinson's disease detection dataset \cite{little2008suitability,tamura2022feature}. The authors used a nonlinear support vector machine to classify subjects into healthy controls or people with Parkinson's disease based on voice measures. After correlation-based filtering and an exhaustive search, they achieved their top accuracy of $0.91$ (95\% confidence interval $[0.87, 0.96]$) with only 4 out of 17 features. For our experiments, we therefore used our cardinality-constrained best subset selection procedure with $k=4$ and a classification threshold at 0.5. We generated 4 cubic splines per feature, with 3 knots distributed along the quartiles of each feature \cite{steyerberg2019clinical}, and applied ``min-max'' scaling. In order to assess the predictive performance, we used a nested, stratified, 3-fold cross-validation in combination with nonparametric bootstrapping. Nested cross-validation consists of an inner loop for hyperparameter tuning, and an outer loop for evaluation. Nonparametric bootstrapping was used to obtain robust 95\% confidence intervals (CIs) that provide a very intuitive interpretation, \ie, that the value of interest will lie in the specified range with a probability of approximately 95\% \cite{hastie2009elements}. In the inner loop, we set our regularization hyperparameter $\lambda$ to one value of the geometric sequence [0.1, 0.02, 0.004], optimized our objective function on 2 of the 3 folds, and used the third fold for validation with the area under the receiver operating characteristic curve (AUC). We repeated this procedure such that each fold was used for validation once, and averaged the validation fold performance over the 3 folds. We repeated these steps for each hyperparameter setting, and then used the best setting to train our predictive model on all 3 folds. In the outer loop, we ran our inner loop on 2 of the 3 folds, and used the third fold for testing with the classification accuracy. We repeated this procedure such that each fold is used for testing once, and averaged the test fold performance over the 3 folds. To obtain robust 95\% CIs, we repeated these steps on 399 nonparametric bootstrap samples, \ie, 399 random samples from the whole data with replacement, and calculated the mean test accuracy 2.5\% and 97.5\% quantiles \cite{davidson2000bootstrap,ledell2015computationally}. Additionally, we registered the selected features.

\vspace{-1mm}

\subsubsection{Results and discussion}
We achieved a mean test accuracy of $0.87$ (95\% CI $[0.84, 0.90]$), similar to \cite{little2008suitability}. In fact, the 95\% CIs between the original authors' and our feature selection approach overlap by a large amount, meaning that there is no statistically significant difference between the approaches ($p>0.05$). While the selected subsets are different, the newly introduced and indeed best performing voice measure in \cite{little2008suitability}, the pitch period entropy, was selected in both approaches. Additionally, our subset included the 5-point and 11-point amplitude perturbation quotients as well as the correlation dimension. In contrast to \cite{little2008suitability}, our best subset selection approach uses a single-step selection procedure and allows for probabilistic risk estimates, which supports transparency.

\subsection{Comparing optimal and greedy cardinality-constrained selection}

\subsubsection{Experimental setup}
For the evaluation on the synthetic data, we solved our mixed-integer conic optimization problem \eqref{eqn:ecc_long} for each run with the cardinality constraint matching the number of informative features on each of the 8 synthetic datasets. We compared the performance of our best subset selection strategy to a greedy feature selection strategy because it is commonly used in prognostic model research \cite{evans2022estimating,moons2015transparent,steyerberg2019clinical} and recommended as a baseline \cite{guyon2003introduction}. As a heuristic, the greedy strategy selected the most valuable features, again matching the number of informative features, based on their univariable association with the outcome (\ie, their parameter magnitude in unregularized univariable logistic regression). This often yields a good, but not necessarily the best subset. The final prognostic model was then built with the filtered features solving eq.~\eqref {eqn:ecc_long} with $\boldsymbol{z} = \boldsymbol{1}$, but without eq.~\eqref{eqn:ecc}. To assess the mean test AUC of both strategies, we performed a nested, stratified, 3-fold cross-validation in combination with nonparametric bootstrapping. Via bootstrapping, we also evaluated the relative number of informative, redundant, and uninformative features selected\footnote{We note that selecting some of the “uninformative” features could in principle still reduce noise and improve prognostic performance \cite{guyon2003introduction}. However, it would still be preferable for interpretability if indeed only the informative feature were recovered.}, as well as how the selected predictors changed with slight perturbations in the data during bootstrapping, \ie, the stability of the selection \cite{breiman2001statistical,steyerberg2019clinical}.

\subsubsection{Results and discussion}

\begin{figure}[tb]
\centering
\subfloat[Cardinality-constrained feature selection\label{fig:fig1}]{%
  \includegraphics[width=\textwidth]{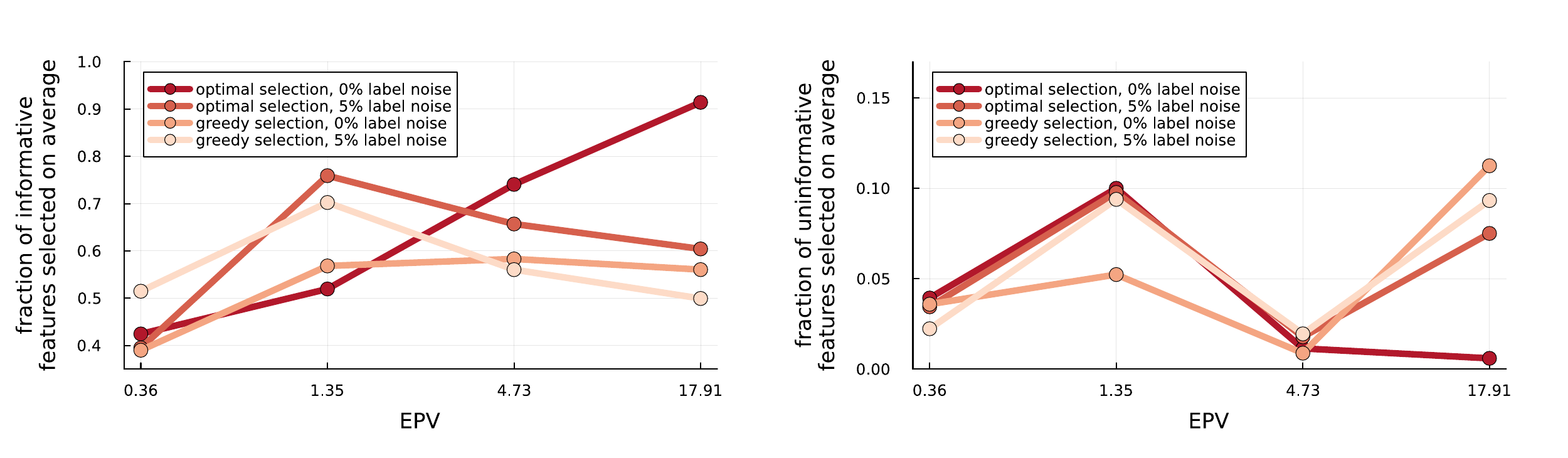}%
}\\[-0.7ex]
\subfloat[Budget-constrained feature selection\label{fig:fig2}]{%
  \includegraphics[width=\textwidth]{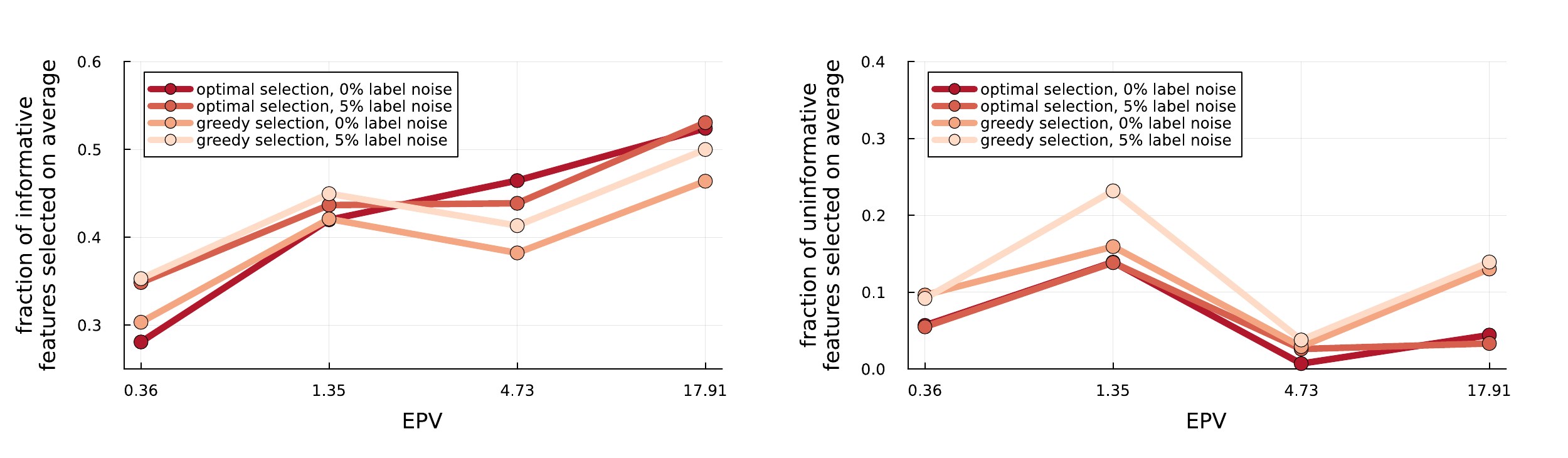}
}
\caption{Feature selection performance with increasing levels of EPVs}
\end{figure}

Mean test AUCs ranged from excellent ($0.80$, 95\% CI $[0.79, 0.98]$) to outstanding ($1.0$, 95\% CI $[1.0, 1.0]$) for optimal selection and from poor ($0.65$, 95\% CI $[0.62, 0.99]$) to outstanding ($1.0$, 95\% CI $[1.0, 1.0]$) for greedy selection. 95\% CIs almost always overlapped to a large extent, meaning that there were generally no statistically significant differences in terms of discrimination between the approaches ($p>0.05$). Without label noise, the fraction of informative features selected on average with optimal selection progressively increased as the EPV increased, from 42\% at $EPV=0.36$ to 91\% at $EPV=17.91$ (see Fig.~\ref{fig:fig1}). The largest fraction of informative features selected on average with greedy selection was 58\% at $EPV=4.73$. While the fraction of redundant features was always less for optimal than for greedy selection, the fraction of uninformative features selected was mostly comparable - the largest difference was observed at $EPV=17.91$, with $<1\%$ on average for optimal selection and 11\% on average for greedy selection. Adding 5\% label noise drastically decreased the largest fraction of informative features selected on average to 76\% for optimal selection at $EPV=1.35$ and increased it to 70\% for greedy selection at $EPV=1.35$. The fraction of redundant features was almost always less for optimal selection than for greedy selection in this case; the fraction of uninformative features selected on average never differed more than 1\% between the methods, but was never $<2\%$. Relatively stable selections across runs were only obtained without label noise for optimal selection at $EPV=17.91$.

Overall, a large fraction of informative features was selected on average at $EPV=17.91$ with optimal selection and without label noise. In this case, a very low fraction of uninformative features was chosen on average and the selection was relatively stable. \emph{Based on our experiments, we therefore recommend our optimal procedure for cardinality-constrained feature selection with an EPV of at least 17.91. Label noise should be avoided at all costs for interpretability, missing label imputations are thus highly questionable in the low-data regime.}

\subsection{Evaluating our cost-sentivive selection algorithm}

\subsubsection{Experimental setup}
For our cost-sensitive best subset selection, we solved the mixed-integer conic optimization problem \eqref{eqn:ecc_long} with eq.~\eqref{eqn:csfs_long} and the budget set to 10 for each run. As a comparator, we used a greedy strategy that selected the most valuable features based on their univariable association with the outcome (\ie, their parameter magnitude in unregularized univariable logistic regression) divided by their individual cost such that the budget was not exceeded. In other words, we chose features with the most bang for the buck. The final prognostic model was built with the filtered features solving eq.~\eqref {eqn:ecc_long} with $\boldsymbol{z} = \boldsymbol{1}$, but without eq.~\eqref{eqn:csfs_long}. Like before, we performed a nested, stratified, 3-fold cross-validation in combination with nonparametric bootstrapping to assess the mean test AUC, the relative number of informative, redundant, and uninformative features selected, as well as the selection stability.

\subsubsection{Results and discussion}

Mean test AUCs ranged from acceptable ($0.75$, 95\% CI $[0.71, 1.0]$) to outstanding ($1.0$, 95\% CI $[0.99, 1.0]$) for optimal selection and from acceptable ($0.76$, 95\% CI $[0.61, 0.94]$) to outstanding ($1.0$, 95\% CI $[0.98, 1.0]$) for greedy selection. 95\% CIs almost always overlapped to a large extent, so there were generally no statistically significant differences in terms of discrimination between the approaches ($p>0.05$). The fraction of informative features selected on average with optimal selection progressively increased as the EPV increased, and was mostly comparable to the fraction selected on average with greedy selection (see Fig.~\ref{fig:fig2}). The fraction of redundant features selected on average was always less for greedy selection than for optimal selection without label noise, with equivocal results when label noise was added. The fraction of uninformative features selected on average was again mostly comparable between the approaches, but in favor of optimal selection especially at $EPV=17.91$. Relatively stable selections across runs were only observed without label noise for optimal selection at $EPV=4.73$ and $EPV=17.91$.

Overall, relatively stable selections were only achieved at $EPV=4.73$ and $EPV=17.91$ with optimal selection and without label noise. A larger fraction of informative features was chosen at $EPV=17.91$, though. In this case, the fraction of uninformative features selected was very low in most runs. \emph{We therefore recommend our optimal cost-sensitive feature selection strategy with an EPV of at least 17.91. As in the cardinality-constrained case, label noise harms interpretability and should be avoided. For this reason, we caution against imputing missing labels even when the sample size is low.}

\section{Conclusion}

Best subset selection is (NP-)hard. Recent advances in mixed-integer programming \cite{bertsimas2021sparse,dedieu2021learning,deza2022safe,tamura2022feature} have made it possible to select the best feature subsets with certificates of optimality, though, even when (sub)set sizes are moderate. We have presented how both cardinality- and budget-constrained best subset selection problems for logistic regression can be formulated from a mixed-integer conic optimization perspective, and how the more general budget constraints are particularly appealing in situations when different costs can be assigned to individual features, for instance when some take longer to measure than others.

Additionally, we have designed a synthetic data generator for clinical prognostic model research. In particular, we have derived sensible parameter settings from the relevant literature for multidimensional pain complaints with information from history-taking and physical examination. We have pointed out conceptual connections of our data synthesis approach to sampling from structural causal models and mentioned how such a prior causal knowledge has been recently used for transformer-based meta-learning \cite{hollmann2022tabpfn}. More broadly, we see that prior knowledge has a huge potential to improve predictive performance when data is limited, as is often the case in the clinical setting. In our view, deep learning models that were pretrained on big data, either real-world or synthetic, and can be adapted to a wide range of downstream tasks, \ie, foundation models, hold a lot of promise in this regard, by capturing effective representations for simpler machine learning models like logistic regression \cite{steinberg2021language,wornow2023shaky}.

Last but not least, we have compared our optimal feature selection approaches to heuristic approaches on diverse synthetic datasets. In line with the literature, we have observed that interpretable results for logistic regression, \ie, solutions with a large fraction of informative features selected, a small fraction of uninformative features selected, and relatively stable selections across runs, were only achieved at an EPV of at least 10 to 15, in our case 17.91 with an optimal selection strategy. We have also observed that a label noise level of already 5\% is detrimental to interpretability, which is why we caution against missing label imputations even when the sample size is low. It is also important to note that different feature sets (for example, with interaction terms or spline transformations), objective functions (for example, with different regularization hyperparameter settings), or machine learning models (for example, nonlinear yet interpretable predictive models like decision trees) would not necessarily yield similar "best" subsets (Sect.~\ref{baseline}), which makes the decision of whether or not a feature is truly relevant even more challenging. Given a predefined feature set and objective function, though, we have shown that cardinality- and budget-constrained best subset selection is indeed possible for logistic regression using mixed-integer conic programming for a sufficiently large EPV without label noise, providing practitioners with transparency and interpretability during predictive model development and validation as commonly required \cite{din2022nrm,ec2021act}. Our optimization problem could be also modified to additionally select a small representative subset of training examples, \ie, a coreset. In the big-data regime, such a reformulation would be beneficial to circumvent training time and memory constraints. This is work in progress, and indeed prior work shows that coresets for our L2-regularized softplus loss can be constructed with high probability by uniform sampling~\cite{curtin2019coresets}. We thank the reviewers for this hint.

\bibliographystyle{splncs04}
\bibliography{mybibliography}

\begin{thebibliography}{10}
\providecommand{\url}[1]{\texttt{#1}}
\providecommand{\urlprefix}{URL }
\providecommand{\doi}[1]{https://doi.org/#1}

\bibitem{abbott2014accuracy}
Abbott, J.H., Kingan, E.M.: Accuracy of physical therapists' prognosis of low
  back pain from the clinical examination: a prospective cohort study. Journal
  of Manual \& Manipulative Therapy  \textbf{22}(3),  154--161 (2014)

\bibitem{aytug2015feature}
Aytug, H.: Feature selection for support vector machines using generalized
  benders decomposition. European Journal of Operational Research
  \textbf{244}(1),  210--218 (2015)

\bibitem{bakker2007spinal}
Bakker, E.W., Verhagen, A.P., Lucas, C., Koning, H.J., Koes, B.W.: Spinal
  mechanical load: a predictor of persistent low back pain? a prospective
  cohort study. European Spine Journal  \textbf{16},  933--941 (2007)

\bibitem{ben2001lectures}
Ben-Tal, A., Nemirovski, A.: Lectures on modern convex optimization: analysis,
  algorithms, and engineering applications. SIAM (2001)

\bibitem{bertsimas2018characterization}
Bertsimas, D., Copenhaver, M.S.: Characterization of the equivalence of
  robustification and regularization in linear and matrix regression. European
  Journal of Operational Research  \textbf{270}(3),  931--942 (2018)

\bibitem{bertsimas2019machine}
Bertsimas, D., Dunn, J.: Machine learning under a modern optimization lens.
  Dynamic Ideas LLC Charlestown, MA (2019)

\bibitem{bertsimas2019robust}
Bertsimas, D., Dunn, J., Pawlowski, C., Zhuo, Y.D.: Robust classification.
  INFORMS Journal on Optimization  \textbf{1}(1),  2--34 (2019)

\bibitem{bertsimas2021sparse}
Bertsimas, D., Pauphilet, J., Van~Parys, B.: Sparse classification: a scalable
  discrete optimization perspective. Machine Learning  \textbf{110},
  3177--3209 (2021)

\bibitem{boyd2004convex}
Boyd, S., Boyd, S.P., Vandenberghe, L.: Convex optimization. Cambridge
  university press (2004)

\bibitem{breiman2001statistical}
Breiman, L.: Statistical modeling: The two cultures (with comments and a
  rejoinder by the author). Statistical science  \textbf{16}(3),  199--231
  (2001)

\bibitem{christodoulou2019systematic}
Christodoulou, E., Ma, J., Collins, G.S., Steyerberg, E.W., Verbakel, J.Y.,
  Van~Calster, B.: A systematic review shows no performance benefit of machine
  learning over logistic regression for clinical prediction models. Journal of
  clinical epidemiology  \textbf{110},  12--22 (2019)

\bibitem{curtin2019coresets}
Curtin, R.R., Im, S., Moseley, B., Pruhs, K., Samadian, A.: On coresets for
  regularized loss minimization. arXiv preprint arXiv:1905.10845  (2019)

\bibitem{davidson2000bootstrap}
Davidson, R., MacKinnon, J.G.: Bootstrap tests: How many bootstraps?
  Econometric Reviews  \textbf{19}(1),  55--68 (2000)

\bibitem{dedieu2021learning}
Dedieu, A., Hazimeh, H., Mazumder, R.: Learning sparse classifiers: Continuous
  and mixed integer optimization perspectives. The Journal of Machine Learning
  Research  \textbf{22}(1),  6008--6054 (2021)

\bibitem{deza2022safe}
Deza, A., Atamturk, A.: Safe screening for logistic regression with l0-l2
  regularization. arXiv preprint arXiv:2202.00467  (2022)

\bibitem{din2022nrm}
{DIN}, {DKE}: {Deutsche Normungsroadmap Künstliche Intelligenz (Ausgabe 2)}.
  \url{https://www.din.de/go/normungsroadmapki/} (2022)

\bibitem{dionne2011five}
Dionne, C.E., Le~Sage, N., Franche, R.L., Dorval, M., Bombardier, C., Deyo,
  R.A.: Five questions predicted long-term, severe, back-related functional
  limitations: evidence from three large prospective studies. Journal of
  clinical epidemiology  \textbf{64}(1),  54--66 (2011)

\bibitem{dunning2017jump}
Dunning, I., Huchette, J., Lubin, M.: Jump: A modeling language for
  mathematical optimization. SIAM review  \textbf{59}(2),  295--320 (2017)

\bibitem{ec2021act}
{European Commission}: {Proposal for a REGULATION OF THE EUROPEAN PARLIAMENT
  AND OF THE COUNCIL LAYING DOWN HARMONISED RULES ON ARTIFICIAL INTELLIGENCE
  (ARTIFICIAL INTELLIGENCE ACT) AND AMENDING CERTAIN UNION LEGISLATIVE ACTS}.
  \url{https://artificialintelligenceact.eu/the-act/} (2021)

\bibitem{evans2022estimating}
Evans, D.W., Rushton, A., Middlebrook, N., Bishop, J., Barbero, M., Patel, J.,
  Falla, D.: Estimating risk of chronic pain and disability following
  musculoskeletal trauma in the united kingdom. JAMA network open
  \textbf{5}(8),  e2228870--e2228870 (2022)

\bibitem{van2021developing}
van~der Gaag, W.H., Chiarotto, A., Heymans, M.W., Enthoven, W.T., van
  Rijckevorsel-Scheele, J., Bierma-Zeinstra, S.M., Bohnen, A.M., Koes, B.W.:
  Developing clinical prediction models for nonrecovery in older patients
  seeking care for back pain: the back complaints in the elders prospective
  cohort study. Pain  \textbf{162}(6), ~1632 (2021)

\bibitem{guyon2003design}
Guyon, I.: Design of experiments of the nips 2003 variable selection benchmark.
  In: NIPS 2003 workshop on feature extraction and feature selection. vol.~253,
  p.~40 (2003)

\bibitem{guyon2003introduction}
Guyon, I., Elisseeff, A.: An introduction to variable and feature selection.
  Journal of machine learning research  \textbf{3}(Mar),  1157--1182 (2003)

\bibitem{guyon2004result}
Guyon, I., Gunn, S., Ben-Hur, A., Dror, G.: Result analysis of the nips 2003
  feature selection challenge. Advances in neural information processing
  systems  \textbf{17} (2004)

\bibitem{guyon2007competitive}
Guyon, I., Li, J., Mader, T., Pletscher, P.A., Schneider, G., Uhr, M.:
  Competitive baseline methods set new standards for the nips 2003 feature
  selection benchmark. Pattern recognition letters  \textbf{28}(12),
  1438--1444 (2007)

\bibitem{hancock2009can}
Hancock, M.J., Maher, C.G., Latimer, J., Herbert, R.D., McAuley, J.H.: Can rate
  of recovery be predicted in patients with acute low back pain? development of
  a clinical prediction rule. European Journal of Pain  \textbf{13}(1),  51--55
  (2009)

\bibitem{harrell2015regression}
Harrell, F.E.: Regression modeling strategies: with applications to linear
  models, logistic regression, and survival analysis. Springer (2015)

\bibitem{hastie2009elements}
Hastie, T., Tibshirani, R., Friedman, J.H., Friedman, J.H.: The elements of
  statistical learning: data mining, inference, and prediction. Springer (2009)

\bibitem{heinze2018variable}
Heinze, G., Wallisch, C., Dunkler, D.: Variable selection--a review and
  recommendations for the practicing statistician. Biometrical journal
  \textbf{60}(3),  431--449 (2018)

\bibitem{hollmann2022tabpfn}
Hollmann, N., M{\"u}ller, S., Eggensperger, K., Hutter, F.: Tabpfn: A
  transformer that solves small tabular classification problems in a second.
  arXiv preprint arXiv:2207.01848  (2022)

\bibitem{kennedy2006eight}
Kennedy, C.A., Haines, T., Beaton, D.E.: Eight predictive factors associated
  with response patterns during physiotherapy for soft tissue shoulder
  disorders were identified. Journal of clinical epidemiology  \textbf{59}(5),
  485--496 (2006)

\bibitem{kuijpers2006clinical}
Kuijpers, T., van~der Windt, D.A., Boeke, A.J.P., Twisk, J.W., Vergouwe, Y.,
  Bouter, L.M., van~der Heijden, G.J.: Clinical prediction rules for the
  prognosis of shoulder pain in general practice. Pain  \textbf{120}(3),
  276--285 (2006)

\bibitem{kuijpers2006prediction}
Kuijpers, T., van~der Windt, D.A., van~der Heijden, G.J., Twisk, J.W.,
  Vergouwe, Y., Bouter, L.M.: A prediction rule for shoulder pain related sick
  leave: a prospective cohort study. BMC musculoskeletal disorders  \textbf{7},
   1--11 (2006)

\bibitem{labbe2019mixed}
Labb{\'e}, M., Mart{\'\i}nez-Merino, L.I., Rodr{\'\i}guez-Ch{\'\i}a, A.M.:
  Mixed integer linear programming for feature selection in support vector
  machine. Discrete Applied Mathematics  \textbf{261},  276--304 (2019)

\bibitem{ledell2015computationally}
LeDell, E., Petersen, M., van~der Laan, M.: Computationally efficient
  confidence intervals for cross-validated area under the roc curve estimates.
  Electronic journal of statistics  \textbf{9}(1), ~1583 (2015)

\bibitem{lee2020mixed}
Lee, I.G., Zhang, Q., Yoon, S.W., Won, D.: A mixed integer linear programming
  support vector machine for cost-effective feature selection. Knowledge-based
  systems  \textbf{203},  106145 (2020)

\bibitem{little2008suitability}
Little, M., McSharry, P., Hunter, E., Spielman, J., Ramig, L.: Suitability of
  dysphonia measurements for telemonitoring of parkinson’s disease. Nature
  Precedings pp.~1--1 (2008)

\bibitem{moons2015transparent}
Moons, K.G., Altman, D.G., Reitsma, J.B., Ioannidis, J.P., Macaskill, P.,
  Steyerberg, E.W., Vickers, A.J., Ransohoff, D.F., Collins, G.S.: Transparent
  reporting of a multivariable prediction model for individual prognosis or
  diagnosis (tripod): explanation and elaboration. Annals of internal medicine
  \textbf{162}(1),  W1--W73 (2015)

\bibitem{moons2019probast}
Moons, K.G., Wolff, R.F., Riley, R.D., Whiting, P.F., Westwood, M., Collins,
  G.S., Reitsma, J.B., Kleijnen, J., Mallett, S.: Probast: a tool to assess
  risk of bias and applicability of prediction model studies: explanation and
  elaboration. Annals of internal medicine  \textbf{170}(1),  W1--W33 (2019)

\bibitem{aps2022mosek}
{MOSEK ApS}: {MOSEK} modeling cookbook (2022)

\bibitem{aps2023mosek}
{MOSEK ApS}: {MOSEK} optimizer {API} for {Python} (2023)

\bibitem{natarajan1995sparse}
Natarajan, B.K.: Sparse approximate solutions to linear systems. SIAM journal
  on computing  \textbf{24}(2),  227--234 (1995)

\bibitem{scheele2013course}
Scheele, J., Enthoven, W.T., Bierma-Zeinstra, S.M., Peul, W.C., van Tulder,
  M.W., Bohnen, A.M., Berger, M.Y., Koes, B.W., Luijsterburg, P.A.: Course and
  prognosis of older back pain patients in general practice: a prospective
  cohort study. PAIN{\textregistered}  \textbf{154}(6),  951--957 (2013)

\bibitem{steinberg2021language}
Steinberg, E., Jung, K., Fries, J.A., Corbin, C.K., Pfohl, S.R., Shah, N.H.:
  Language models are an effective representation learning technique for
  electronic health record data. Journal of biomedical informatics
  \textbf{113},  103637 (2021)

\bibitem{steyerberg2019clinical}
Steyerberg, E.W.: Clinical prediction models: a practical approach to
  development, validation, and updating. Springer (2019)

\bibitem{tamura2022feature}
Tamura, R., Takano, Y., Miyashiro, R.: Feature subset selection for kernel svm
  classification via mixed-integer optimization. arXiv preprint
  arXiv:2205.14325  (2022)

\bibitem{tibshirani1996regression}
Tibshirani, R.: Regression shrinkage and selection via the lasso. Journal of
  the Royal Statistical Society: Series B (Methodological)  \textbf{58}(1),
  267--288 (1996)

\bibitem{tillmann2021cardinality}
Tillmann, A.M., Bienstock, D., Lodi, A., Schwartz, A.: Cardinality
  minimization, constraints, and regularization: a survey. arXiv preprint
  arXiv:2106.09606  (2021)

\bibitem{wippert2017development}
Wippert, P.M., Puschmann, A.K., Drie{\ss}lein, D., Arampatzis, A., Banzer, W.,
  Beck, H., Schiltenwolf, M., Schmidt, H., Schneider, C., Mayer, F.:
  Development of a risk stratification and prevention index for stratified care
  in chronic low back pain. focus: yellow flags (mispex network). Pain reports
  \textbf{2}(6) (2017)

\bibitem{wornow2023shaky}
Wornow, M., Xu, Y., Thapa, R., Patel, B., Steinberg, E., Fleming, S., Pfeffer,
  M.A., Fries, J., Shah, N.H.: The shaky foundations of clinical foundation
  models: A survey of large language models and foundation models for emrs.
  arXiv preprint arXiv:2303.12961  (2023)

\end{thebibliography}

\newpage
\section{Appendix}

\begin{figure}
\centering
\subfloat[$M=82,\,N=53,\,EPV=0.36$\label{fig:fig3}]{%
  \includegraphics[width=0.84\textwidth]{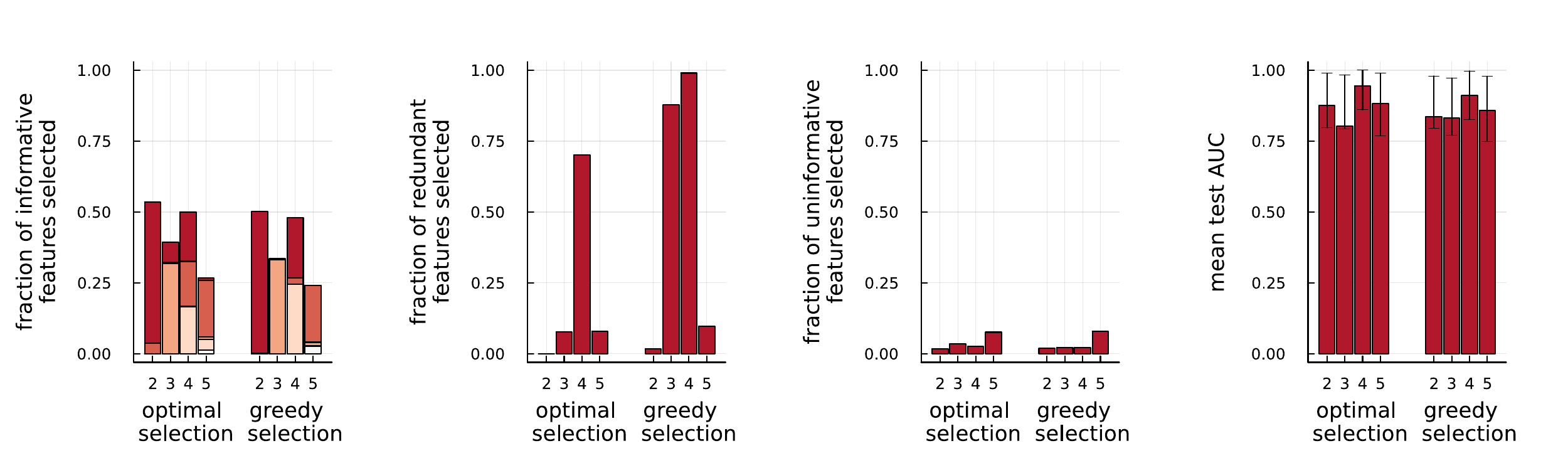}%
}\\[-0.7ex]
\subfloat[$M=82,\,N=14,\,EPV=1.35$\label{fig:fig4}]{%
  \includegraphics[width=0.84\textwidth]{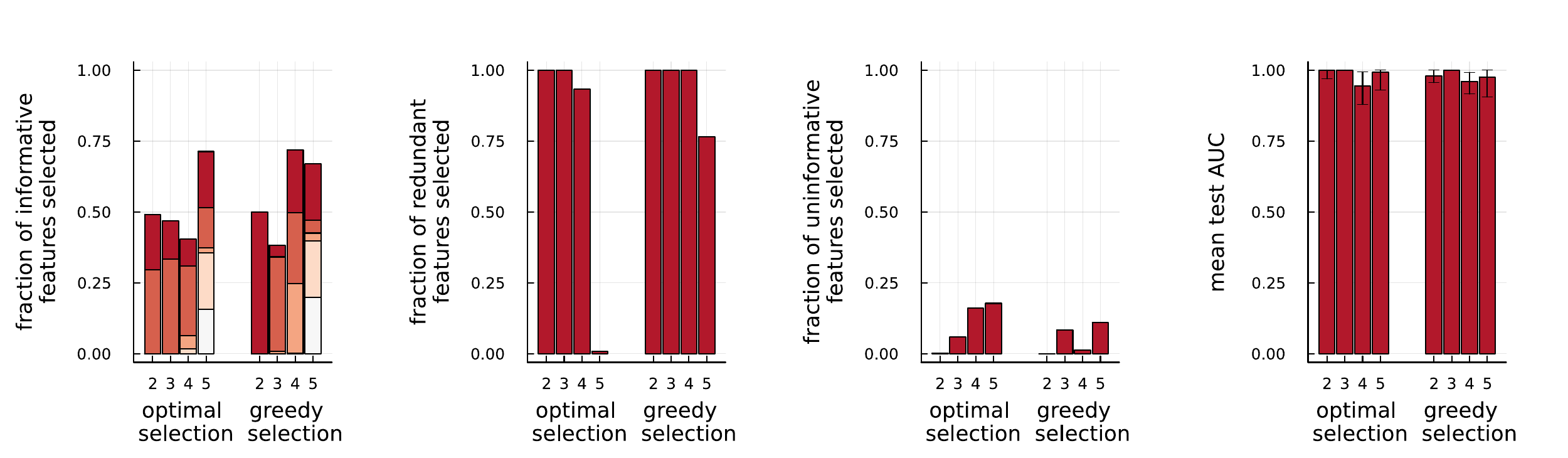}
}\\[-0.7ex]
\subfloat[$M=1090,\,N=53,\,EPV=4.73$\label{fig:fig5}]{%
  \includegraphics[width=0.84\textwidth]{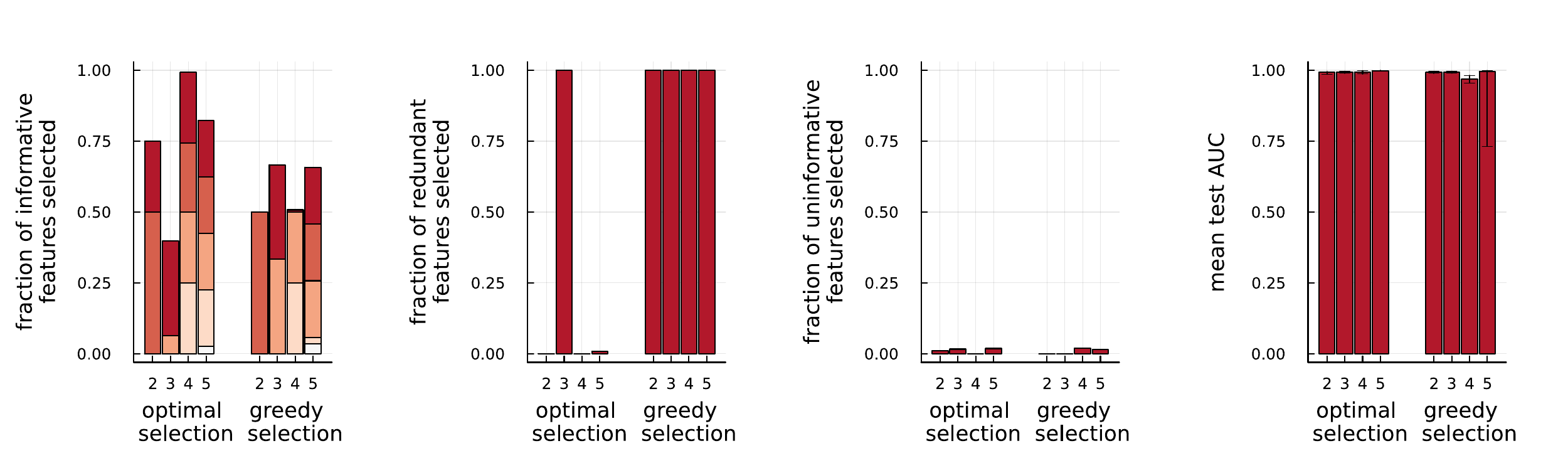}%
}\\[-0.7ex]
\subfloat[$M=1090,\,N=14,\,EPV=17.91$\label{fig:fig6}]{%
  \includegraphics[width=0.84\textwidth]{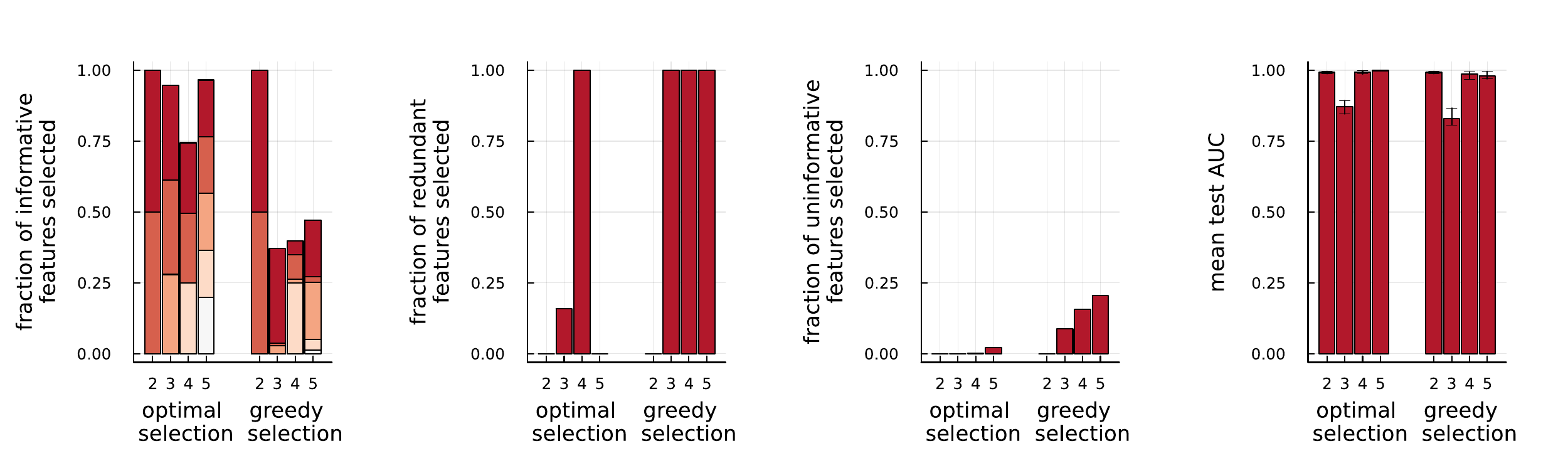}%
}
\caption{Cardinality-constrained feature selection, label noise level 0\%}
\label{fig7}
\end{figure}

\begin{figure}
\centering
\subfloat[$M=82,\,N=53,\,EPV=0.36$\label{fig:fig8}]{%
  \includegraphics[width=0.84\textwidth]{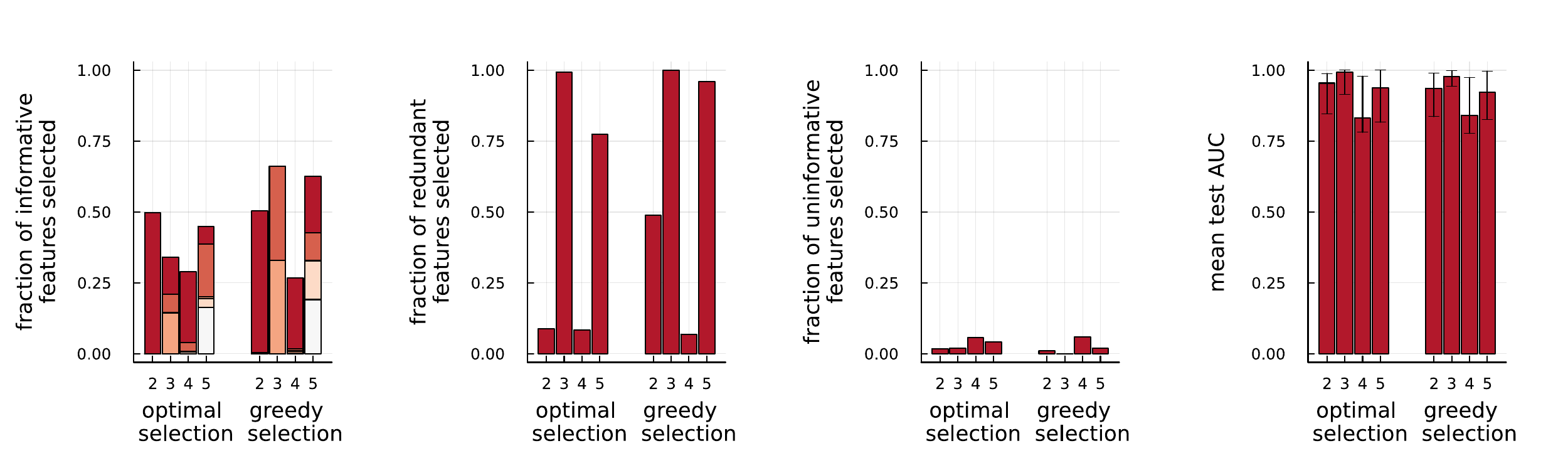}%
}\\[-0.7ex]
\subfloat[$M=82,\,N=14,\,EPV=1.35$\label{fig:fig9}]{%
  \includegraphics[width=0.84\textwidth]{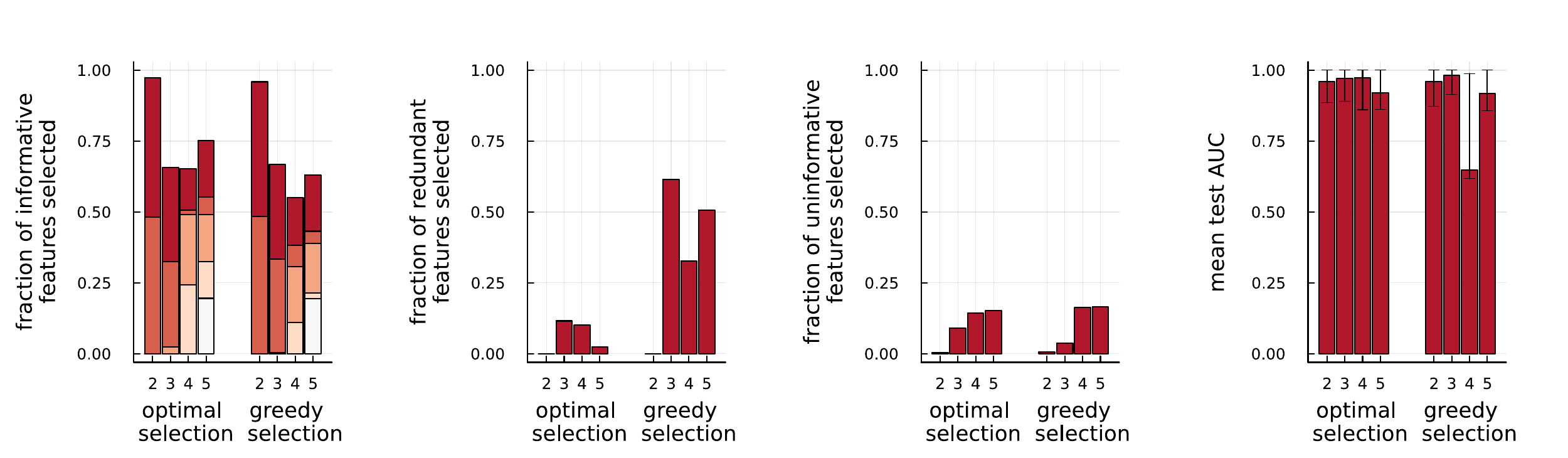}
}\\[-0.7ex]
\subfloat[$M=1090,\,N=53,\,EPV=4.73$\label{fig:fig10}]{%
  \includegraphics[width=0.84\textwidth]{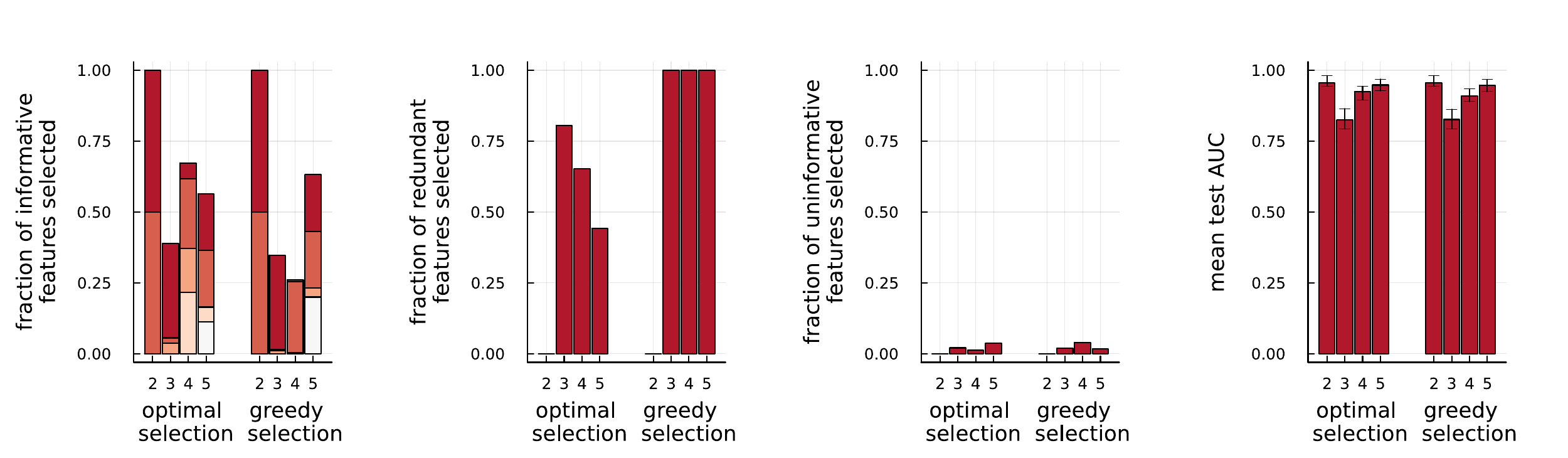}%
}\\[-0.7ex]
\subfloat[$M=1090,\,N=14,\,EPV=17.91$\label{fig:fig11}]{%
  \includegraphics[width=0.84\textwidth]{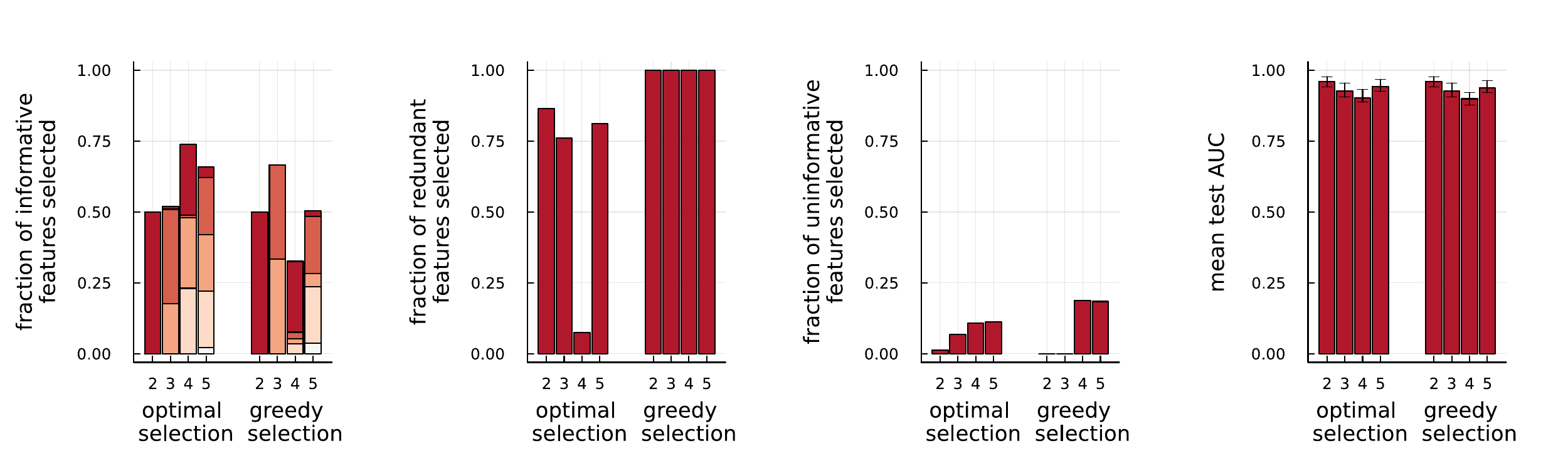}%
}
\caption{Cardinality-constrained feature selection, label noise level 5\%}
\label{fig12}
\end{figure}

\begin{figure}
\centering
\subfloat[$M=82,\,N=53,\,EPV=0.36$\label{fig:fig13}]{%
  \includegraphics[width=0.84\textwidth]{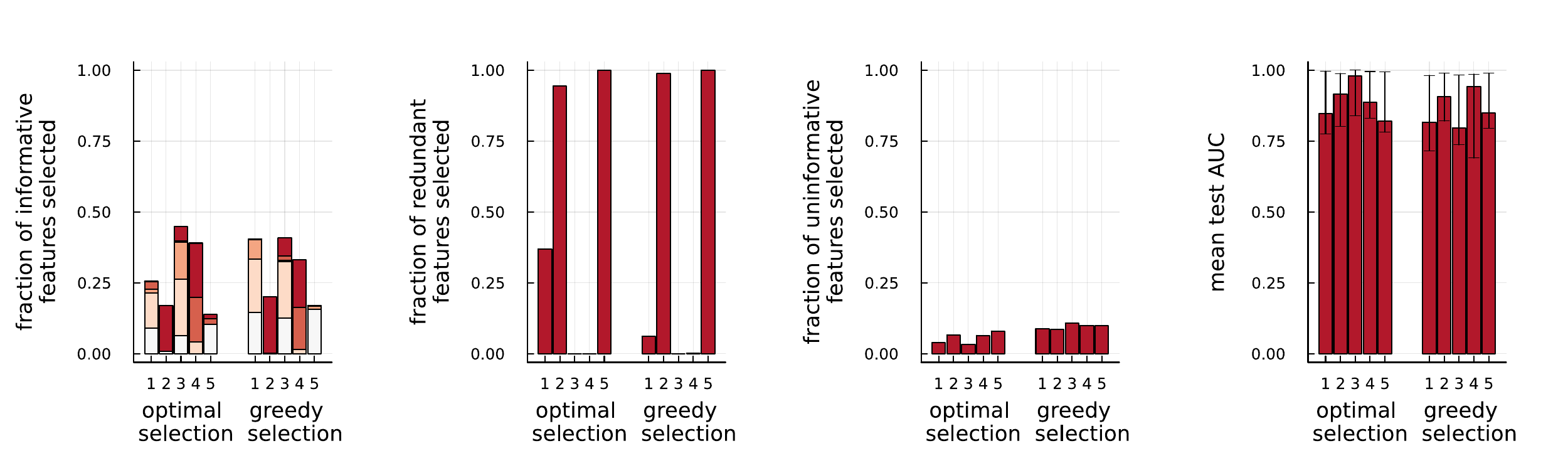}%
}\\[-0.7ex]
\subfloat[$M=82,\,N=14,\,EPV=1.35$\label{fig:fig14}]{%
  \includegraphics[width=0.84\textwidth]{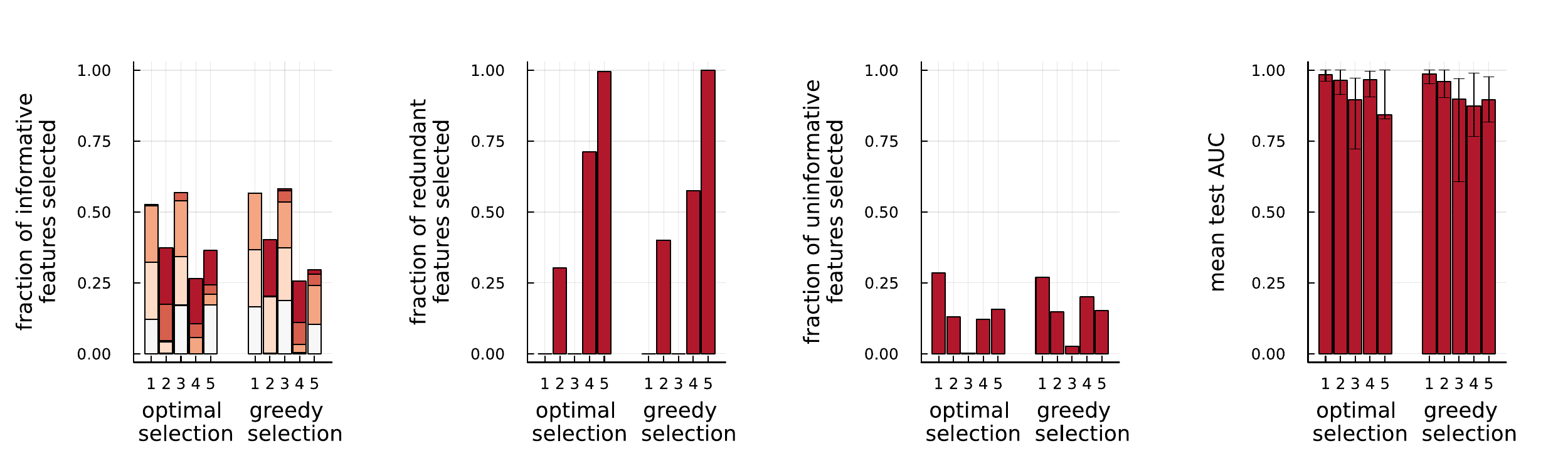}
}\\[-0.7ex]
\subfloat[$M=1090,\,N=53,\,EPV=4.73$\label{fig:fig15}]{%
  \includegraphics[width=0.84\textwidth]{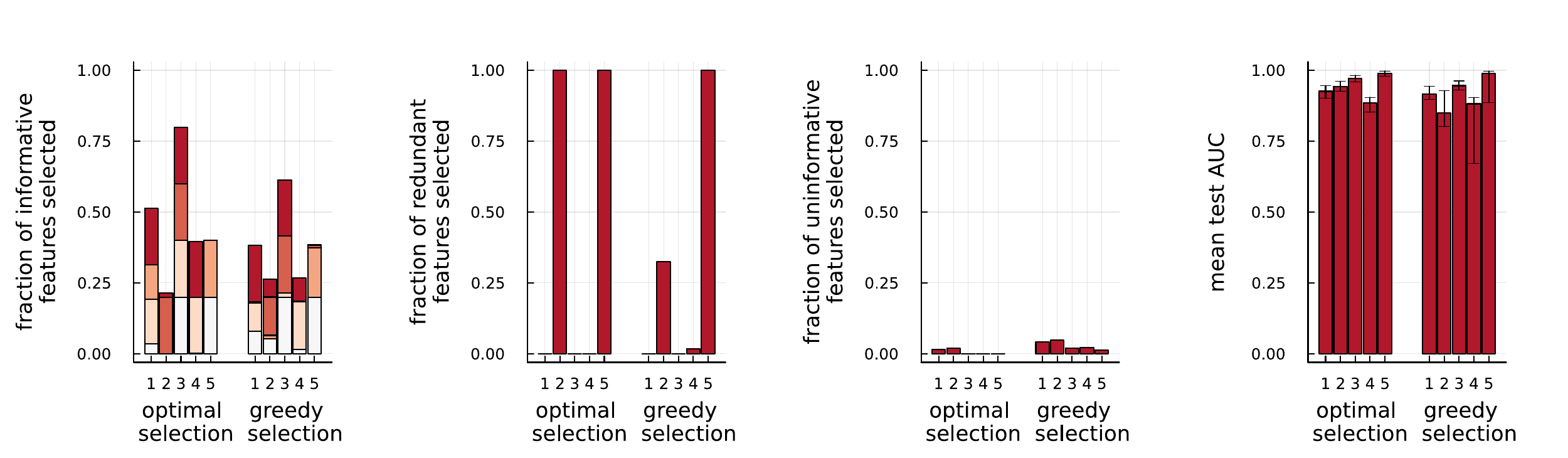}%
}\\[-0.7ex]
\subfloat[$M=1090,\,N=14,\,EPV=17.91$\label{fig:fig16}]{%
  \includegraphics[width=0.84\textwidth]{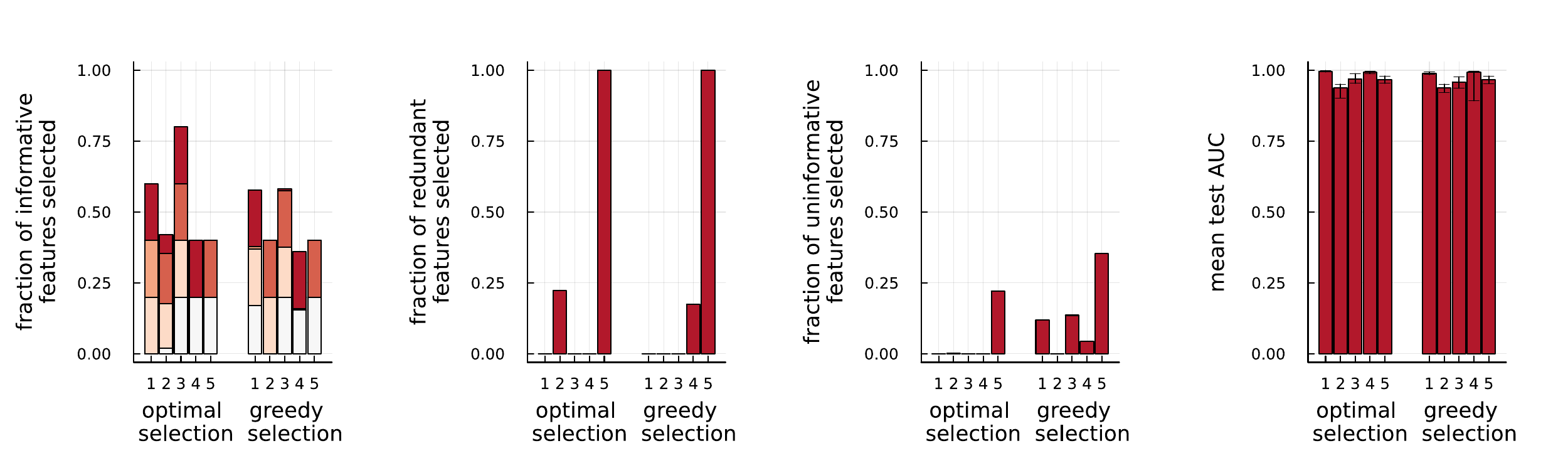}%
}
\caption{Budget-constrained feature selection, label noise level 0\%}
\label{fig17}
\end{figure}

\begin{figure}
\centering
\subfloat[$M=82,\,N=53,\,EPV=0.36$\label{fig:fig18}]{%
  \includegraphics[width=0.84\textwidth]{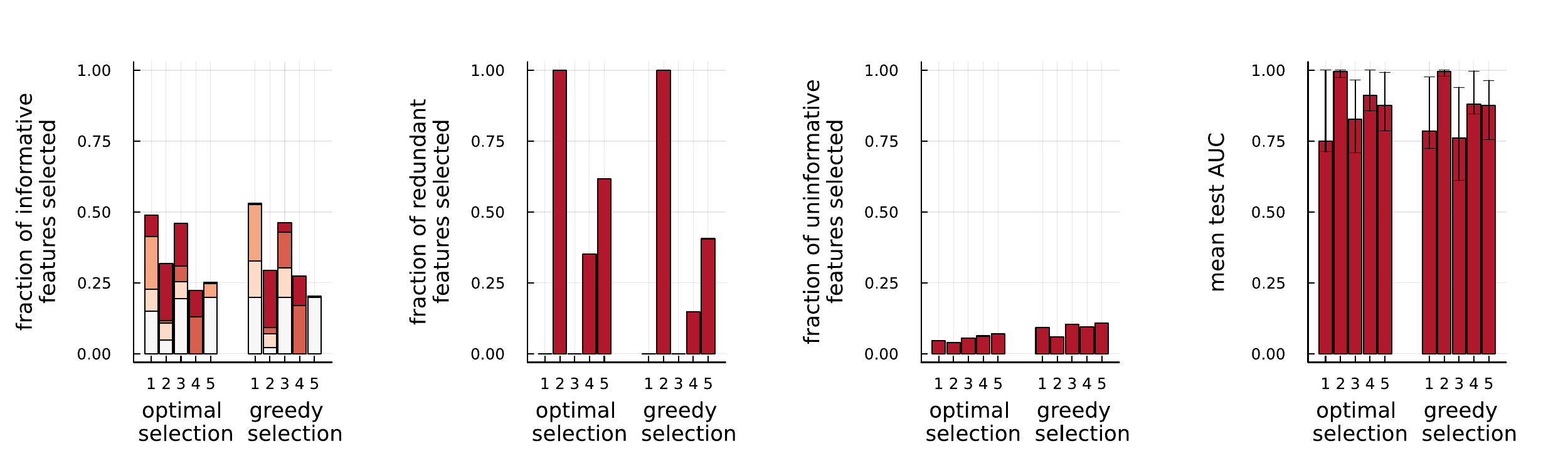}%
}\\[-0.7ex]
\subfloat[$M=82,\,N=14,\,EPV=1.35$\label{fig:fig19}]{%
  \includegraphics[width=0.84\textwidth]{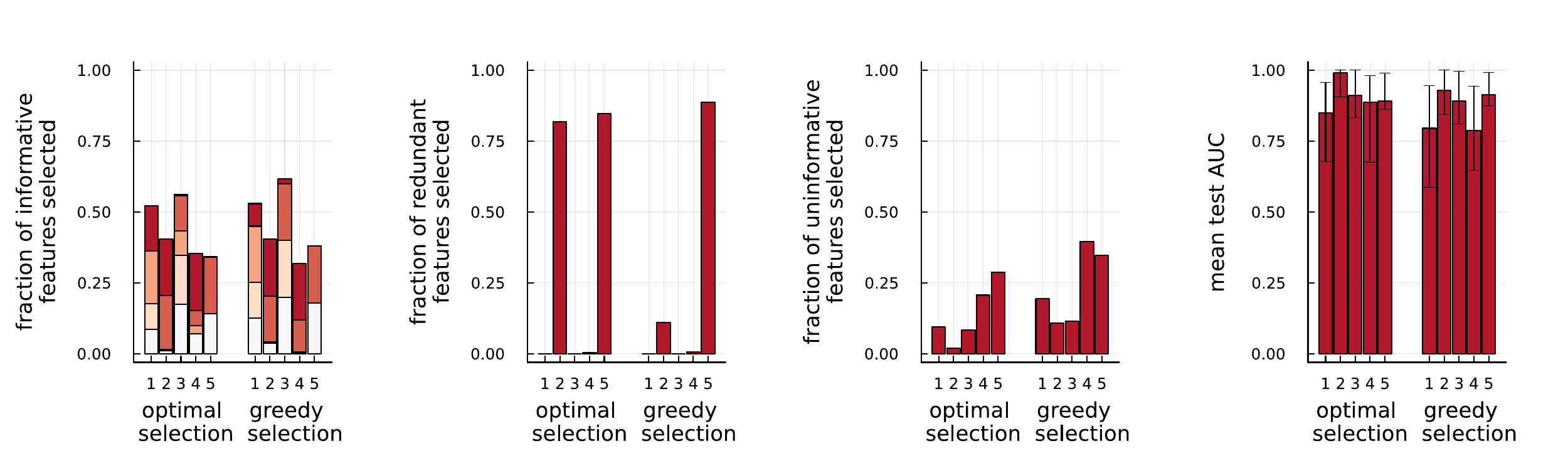}
}\\[-0.7ex]
\subfloat[$M=1090,\,N=53,\,EPV=4.73$\label{fig:fig20}]{%
  \includegraphics[width=0.84\textwidth]{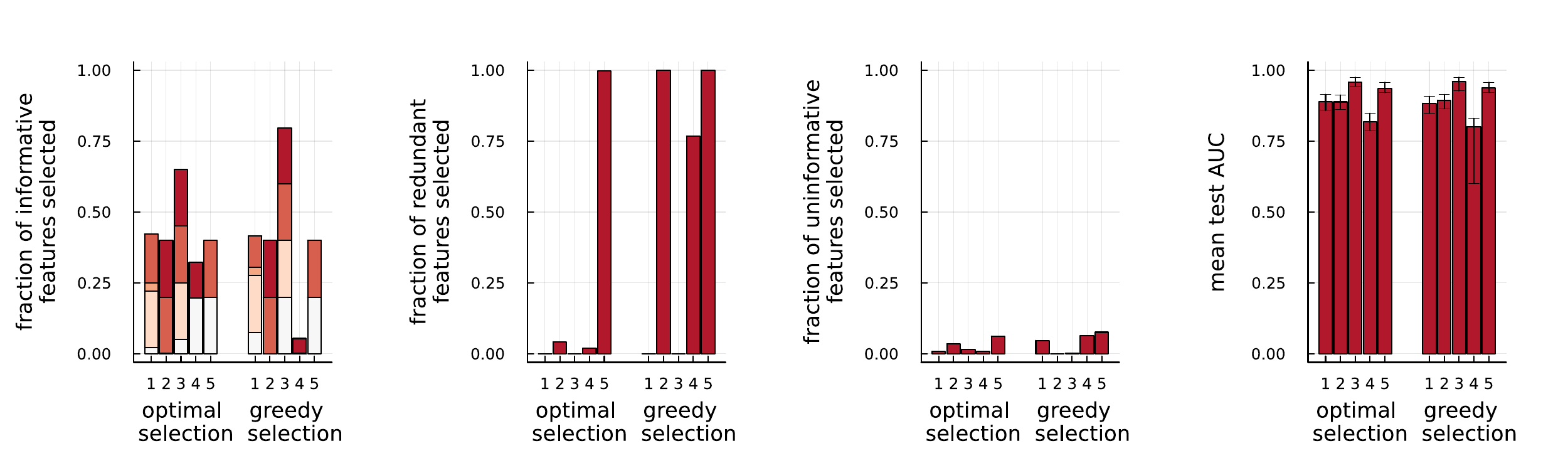}%
}\\[-0.7ex]
\subfloat[$M=1090,\,N=14,\,EPV=17.91$\label{fig:fig21}]{%
  \includegraphics[width=0.84\textwidth]{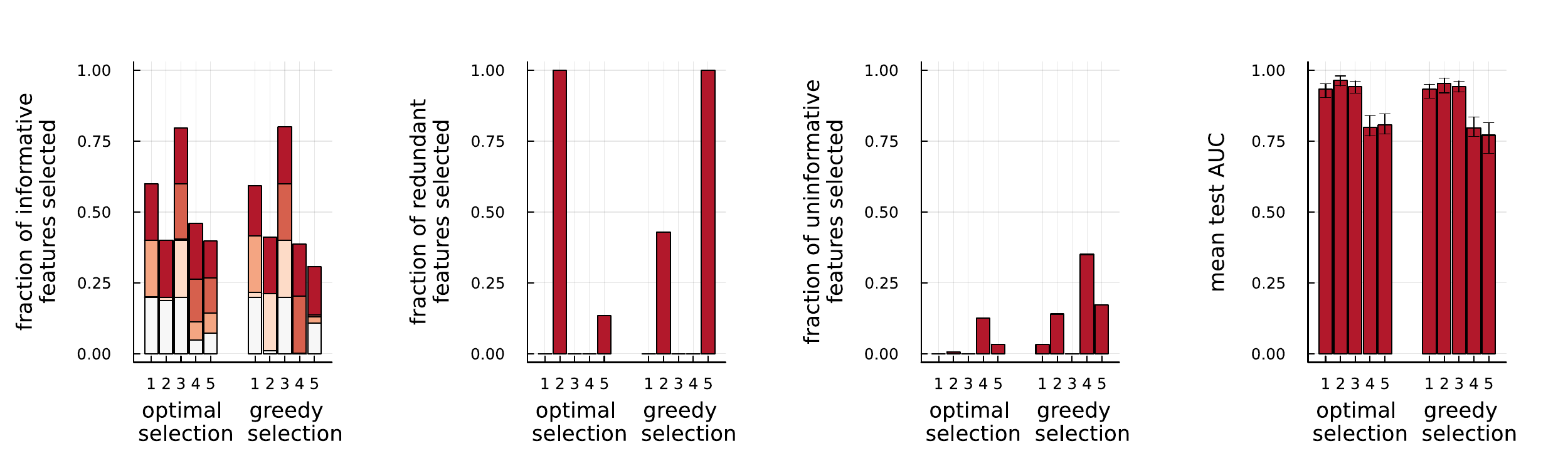}%
}
\caption{Budget-constrained feature selection, label noise level 5\%}
\label{fig22}
\end{figure}

Figures \ref{fig7} to \ref{fig22} show the detailed experimental results. Each informative feature is represented by a different hue, error bars represent 95\% CIs.

\end{document}